\documentclass{article}

\usepackage{times}
\usepackage{graphicx} 
\usepackage{subfigure}

\usepackage{natbib}

\usepackage{algorithm}
\usepackage{algorithmic}
\usepackage[algo2e,linesnumbered,lined,boxed,vlined]{algorithm2e}

\usepackage{amsfonts}
\usepackage{amssymb}

\usepackage{hyperref}

\usepackage[accepted]{icml2016}

\usepackage{comment}
\usepackage{amsthm}
\usepackage{mathtools}
\usepackage{tabularx}
\usepackage{multirow}

\def\b{\ensuremath\boldsymbol}

\usepackage{lastpage}
\usepackage{fancyhdr}
\usepackage{url}
\icmltitlerunning{}

\begin{document}

\twocolumn[
\icmltitle{Probabilistic Classification by Density Estimation \\Using Gaussian Mixture Model and Masked Autoregressive Flow}

\icmlauthor{Benyamin Ghojogh*}{bghojogh@uwaterloo.ca}
\icmladdress{Department of Electrical and Computer Engineering, University of Waterloo, Waterloo, ON, Canada}
\icmlauthor{Milad Amir Toutounchian*}{mt3393@drexel.edu}
\icmladdress{College of Computing \& Informatics, Drexel University, Philadelphia, Pennsylvania, USA
\\ \\
*The two authors contributed equally to this work.
}


\vskip 0.3in
]

\begin{abstract}
Density estimation, which estimates the distribution of data, is an important category of probabilistic machine learning. A family of density estimators is mixture models, such as Gaussian Mixture Model (GMM) by expectation maximization. Another family of density estimators is the generative models which generate data from input latent variables. One of the generative models is the Masked Autoregressive Flow (MAF) which makes use of normalizing flows and autoregressive networks. In this paper, we use the density estimators for classification, although they are often used for estimating the distribution of data. We model the likelihood of classes of data by density estimation, specifically using GMM and MAF. The proposed classifiers outperform simpler classifiers such as linear discriminant analysis which model the likelihood using only a single Gaussian distribution. This work opens the research door for proposing other probabilistic classifiers based on joint density estimation. 
\end{abstract}


\section{Introduction}\label{section_introduction}

Probabilistic machine learning makes use of the theory of probability for different tasks such as classification, regression, clustering, dimensionality reduction, and data generation. 
Some of the fundamental methods for probabilistic classification are Linear Discriminant Analysis (LDA) \cite{ghojogh2019linear}, Bayes, and naive Bayes classifiers \cite{murphy2012machine}.  
A broad category of probabilistic machine learning is the generative models which generate data distribution from some input latent variable. 
One of the most fundamental generative models is factor analysis \cite{fruchter1954introduction,ghojogh2023factor} whose special case is probabilistic principal component analysis \cite{roweis1997algorithms,tipping1999probabilistic}.

By the development of neural networks, they were used as probabilistic generative models \cite{durr2020probabilistic}.
One of them is variational autoencoder \cite{kingma2014auto} which models variational inference in a neural network. 
Another generative model is generative adversarial network \cite{goodfellow2014generative} which has also been implemented as an autoencoder \cite{makhzani2018unsupervised}.
Another successful family of the generative models is the diffusion models \cite{ho2020denoising}. 
Finally, one other category of generative models is normalizing flows \cite{rezende2015variational} which transform a distribution of the latent noise to the desired distribution of data. 
There are various variants of normalizing flows \cite{kobyzev2020normalizing} some of which are Masked Autoencoder for Distribution Estimation (MADE) \cite{germain2015made}, real-valued Non-Volume Preserving (real NVP) \cite{dinh2016density}, Masked Autoregressive Flow (MAF) \cite{papamakarios2017masked}, and triangular neural network \cite{li2020triangular}.

In addition to the generative models, some probabilistic methods are used for density estimation. These methods fit a probability density function to a sample of data. 
One of these methods is the Gaussian Mixture Model (GMM) which models the distribution as a mixture of normal distributions. The mixture models can be trained using various methods; one of its mostly used approaches is the expectation maximization \cite{ghojogh2019fitting}. 
Other approaches, such as using Riemannian optimization \cite{hosseini2020alternative} and kernel density estimation \cite{chen2017tutorial}, also exist for density estimation and fitting mixture models.
Note that the generative models can also be analyzed as density estimators because they estimate the distribution of data from the distribution of latent variable. 

On the one hand, the generative models, as special cases of density estimators, are mostly used for density estimation, data generation, or dimensionality reduction \cite{makhzani2018unsupervised}. 
On the other hand, the mixture models are used for density estimation. 
In this paper, we propose to use the density estimators, including both generative models and mixture models, for the task of classification. 
We use MAF and GMM as two examples of the generative models and the mixture models, respectively. The idea of using density estimation models for classification can be extended to using other density estimators. 

The remainder of this paper is organized as follows. Sections \ref{section_GMM_classifier} and \ref{section_MAF_classifier} introduce the theory of GMM and MAF and explain the proposed GMM and MAF classifiers, respectively. 
Simulations on toy and real-world data are reported in Section \ref{section_simulations}. Finally, Section \ref{section_conclusion} concludes the paper and enumerates the future directions. 

\section{Gaussian Mixture Model Classifier}\label{section_GMM_classifier}

\subsection{LDA and Bayes' Classifiers}

Consider a dataset of size $n$ and dimensionality $d$, i.e., $\{(\b{x}_i, y_i)\}_{i=1}^n$, where $\b{x}_i \in \mathbb{R}^d$ and $y_i \in \mathbb{R}$ are the $i$-th data point and its class label, respectively. 
Let $|\mathcal{C}|$ denote the number of classes and $\mathcal{C}_j$ denote the $j$-th class. 
According to the Bayes' rule, 
\begin{align}
&\mathbb{P}(\b{x} \in \mathcal{C}_j \,|\, X=\b{x}) = \frac{\mathbb{P}(X=\b{x} \,|\, \b{x} \in \mathcal{C}_j)\, \mathbb{P}(\b{x} \in \mathcal{C}_j)}{\mathbb{P}(X=\b{x})} \nonumber \\
&~~~~~~~~~~~~~~~~ = \frac{\mathbb{P}(X=\b{x} \,|\, \b{x} \in \mathcal{C}_j)\, \mathbb{P}(\b{x} \in \mathcal{C}_j)}{\sum_{k=1}^{|\mathcal{C}|} \mathbb{P}(X=\b{x} \,|\, \b{x} \in \mathcal{C}_k)\, \mathbb{P}(\b{x} \in \mathcal{C}_k)}, \label{Bayes_rule}
\end{align}
$\forall j \in \{1, \dots, |\mathcal{C}|\}$, where $X$ denotes the random variable of data.
The probabilities $\mathbb{P}(\b{x} \in \mathcal{C}_j \,|\, X=\b{x})$, $\mathbb{P}(X=\b{x} \,|\, \b{x} \in \mathcal{C}_j)$, and $\mathbb{P}(\b{x} \in \mathcal{C}_j)$ are the posterior, likelihood, and the class prior, respectively.  

LDA, Quadratic Discriminant Analysis (QDA), and Bayes' classifiers set the decision boundary of classification where the posterior of two adjacent classes are equal \cite{ghojogh2019linear}:
\begin{align}
\mathbb{P}(\b{x} \in \mathcal{C}_1 \,|\, X=\b{x}) = \mathbb{P}(\b{x} \in \mathcal{C}_2 \,|\, X=\b{x}).
\end{align}
In other words, the class which maximizes the posterior of a data instance determines its class:
\begin{align}\label{equation_maximize_posterior}
&\underset{j \in \{1, \dots, |\mathcal{C}|\}}{\text{maximize}}\quad \mathbb{P}(\b{x} \in \mathcal{C}_j \,|\, X=\b{x}), 
\end{align}
which is equivalent to maximizing the scaled multiplication of likelihood and class prior, according to Eq. (\ref{Bayes_rule}):
\begin{align}\label{equation_maximize_likelihood_times_prior}
&\underset{j \in \{1, \dots, |\mathcal{C}|\}}{\text{maximize}}\quad \mathbb{P}(X=\b{x} \,|\, \b{x} \in \mathcal{C}_j)\, \mathbb{P}(\b{x} \in \mathcal{C}_j).
\end{align}
The class prior $\mathbb{P}(\b{x} \in \mathcal{C}_j)$ can be estimated by the fraction of the cardinality of class over the total sample size.

Eq. (\ref{equation_maximize_posterior}) or (\ref{equation_maximize_likelihood_times_prior}) is the optimization problem in the Bayes' classifier \cite{murphy2012machine}.
The Bayes' classifier is an optimal classifier
(see {\citep[Chapter 6]{mitchell1997machine}} and \cite{zhang2004optimality}).
LDA and QDA consider Gaussian distributions for the likelihoods in Eq. (\ref{equation_maximize_likelihood_times_prior}):
\begin{equation}\label{equation_Gaussian_PDF}
\begin{aligned}
& \mathbb{P}(X=\b{x} \,|\, \b{x} \in \mathcal{C}_j) \\
&= \frac{1}{\sqrt{(2\pi)^d \text{det}(\b{\Sigma}_j)}} \exp\! \Big(\!-\frac{(\b{x}-\b{\mu}_j)^\top \b{\Sigma}_j^{-1} (\b{x}-\b{\mu}_j)}{2} \Big),
\end{aligned}
\end{equation}
where $\b{\mu}_j \in \mathbb{R}^d$ and $\b{\Sigma}_j \in \mathbb{R}^{d \times d}$ are the mean and covariance of the $j$-th class and $\text{det}(.)$ denotes the determinant of matrix.
An additional assumption in LDA, which makes the decision boundary linear, is considering equal covariance matrices for the classes \cite{ghojogh2019linear}. 

\subsection{Gaussian Mixture Model}

\subsubsection{GMM for Density Estimation}

The mixture distribution is a weighted summation of $k$ distributions $\{g_j(\b{x}; \Theta_j)\}_{j=1}^k$ where the weights $\{w_1, \dots, w_k\}$ sum to one. The mixture distribution is formulated as \cite{ghojogh2019fitting}:
\begin{equation}\label{equation_mixture}
\begin{aligned}
& f(\b{x}; \Theta_1, \dots, \Theta_k) = \sum_{j=1}^k w_j\, g_j(\b{x}; \Theta_j), \\
& \text{subject to} ~~~~ \sum_{j=1}^k w_j = 1.
\end{aligned}
\end{equation}

In GMM, the distributions $\{g_j(\b{x}; \Theta_j)\}_{j=1}^k$ are the Gaussian distributions $\{g_j(\b{x}; \b{\mu}_j, \b{\Sigma}_j)\}_{j=1}^k$ whose parameters are the means and covariances. Every distribution $g_j(\b{x}; \b{\mu}_j, \b{\Sigma}_j)$ is formulated as in Eq. (\ref{equation_Gaussian_PDF}). 
GMM is effective because the distributions in the natural life are often Gaussian for the central limit theorem \cite{hazewinkel2001central}.

The optimization problem (\ref{equation_mixture}) can be solved using the Expectation Maximization (EM) algorithm where the mixing probabilities $\{w_j\}_{j=1}^k$, means, and covariances are updated iteratively as:
\begin{align}
& \widehat{w}_j = \frac{\sum_{i=1}^n \widehat{\gamma}_{i,j}}{\sum_{i=1}^n \sum_{z=1}^k \widehat{\gamma}_{i,z}}, \label{equation_w_multiMixture} \\
& \widehat{\b{\mu}}_j = \frac{\sum_{i=1}^n \widehat{\gamma}_{i,j}\, \b{x}_i}{\sum_{i=1}^n \widehat{\gamma}_{i,j}}, \label{equation_mu_multiMixture} \\
& \widehat{\b{\Sigma}}_j = \frac{\sum_{i=1}^n \widehat{\gamma}_{i,j} (\b{x}_i - \widehat{\b{\mu}}_j)(\b{x}_i - \widehat{\b{\mu}}_j)^\top}{\sum_{i=1}^n \widehat{\gamma}_{i,j}}, \label{equation_Sigma_multiMixture}
\end{align}
where:
\begin{align}
& \widehat{\gamma}_{i,j} = \frac{\widehat{w}_j\, g_j(\b{x}_i; \widehat{\b{\mu}}_j, \widehat{\b{\Sigma}}_j)}{\sum_{z=1}^k \widehat{w}_{z}\, g_{z}(\b{x}_i; \widehat{\b{\mu}}_{z}, \widehat{\b{\Sigma}}_{z})}. \label{equation_gamma_multiMixture} 
\end{align}
Refer to \cite{ghojogh2019fitting} for the derivation of these equations. 
The EM algorithm for training GMM is in Algorithm \ref{algorithm_GMM}.

\SetAlCapSkip{0.5em}
\IncMargin{0.8em}
\begin{algorithm2e}[!t]
\DontPrintSemicolon
	\textbf{Initialize} $\widehat{\b{\mu}}_1, \dots, \widehat{\b{\mu}}_k$, $\widehat{\b{\Sigma}}_1, \dots, \widehat{\b{\Sigma}}_k$, $\widehat{w}_1, \dots, \widehat{w}_k$\;
	\While{not converged}{
	    \textit{// E-step in EM:}\;
	    \For{$i$ from $1$ to $n$}{
	        \For{$j$ from $1$ to $k$}{
                $\widehat{\gamma}_{i,j}  \gets $ Eq. (\ref{equation_gamma_multiMixture})\;
            }
        }
        \textit{// M-step in EM:}\;
        \For{$j$ from $1$ to $k$}{
    	    $\widehat{\b{\mu}}_j  \gets $ Eq. (\ref{equation_mu_multiMixture})\;
                $\widehat{\b{\Sigma}}_j  \gets $ Eq. (\ref{equation_Sigma_multiMixture})\;
    	    $\widehat{w}_j \gets $ Eq. (\ref{equation_w_multiMixture})\;
	    }
	    \textit{// Check convergence:}\;
	    Compare $\widehat{\b{\mu}}_1, \dots, \widehat{\b{\mu}}_k$, $\widehat{\b{\Sigma}}_1, \dots, \widehat{\b{\Sigma}}_k$, and $\widehat{w}_1, \dots, \widehat{w}_k$ with their values in the previous iteration
	}
\caption{GMM Training using the EM Algorithm}\label{algorithm_GMM}
\end{algorithm2e}
\DecMargin{0.8em}

\subsubsection{GMM for Classification}

GMM can be used for classification in addition to density estimation. 
Recall the optimization problem (\ref{equation_maximize_likelihood_times_prior}) in the Bayes' classifier for determining the class of a data instance. As explained before, LDA uses Gaussian distribution for the likelihood of each class in this optimization problem.  
This works well when every class actually has a uni-modal Gaussian distribution.
However, LDA will perform poorly when every class either has a multi-modal Gaussian distribution or a completely different distribution from the Gaussian density \cite{sugiyama2007dimensionality}. 

Our proposed GMM classifier fits a mixture of Gaussian distributions, rather than only one Gaussian distribution, to the likelihood of every class:
\begin{align}
\mathbb{P}(X=\b{x} \,|\, \b{x} \in \mathcal{C}_z) \approx \sum_{j=1}^k w_j^{z}\, g_j^{z}(\b{x}; \b{\mu}^{z}_j, \b{\Sigma}_j^{z}),
\end{align}
where the superscript $z$ shows the class index.
Then, the maximum posterior among classes determines the class of a data instance $\b{x}$:
\begin{align}\label{equation_GMM_classifier}
&\underset{z \in \{1, \dots, |\mathcal{C}|\}}{\text{maximize}}\quad \mathbb{P}(\b{x} \in \mathcal{C}_z) \sum_{j=1}^k w_j^{z}\, g_j^{z}(\b{x}; \b{\mu}^{z}_j, \b{\Sigma}_j^{z}).
\end{align}
This improves the performance of classification because of the following explanation. On the one hand, if every class has a multi-modal Gaussian distribution, it is enough to set the number of Gaussians, $k$, in the mixture to the number of modes. 
On the other hand, if the distribution of a class is strange and complicated, a sufficiently large $k$ can fit a good mixture of Gaussians to the class. 
Note that the hyper-parameter $k$ can be determined by validation. 

\section{Masked Autoregressive Flow Classifier}\label{section_MAF_classifier}

\subsection{Normalizing Flows}

Consider two random variables $s$ and $u$, where we want to transform the latent variable $u$ to $s$ by a function $g: u \mapsto s$ as:
\begin{align}\label{equation_function_g}
s = g(u),
\end{align}
in a way that the Probability Density Function (PDF) is preserved in this transformation.
As the PDF is continuous, the following should hold for it to be preserved:
\begin{align}\label{equation_Px_dx_Pg_dg}
\mathbb{P}_s(s)\, |ds| = \mathbb{P}_u(u)\, |du|,
\end{align}
where $|.|$ denotes the absolute value and $dx$ and $du$ are differentials of $x$ and $u$, respectively.
According to Eqs. (\ref{equation_function_g}) and (\ref{equation_Px_dx_Pg_dg}), we have:
\begin{align}
\mathbb{P}_s(s) &= \mathbb{P}_u(u)\, \Big|\frac{du}{ds}\Big| = \mathbb{P}_u(u)\, \Big|\frac{ds}{du}\Big|^{-1} \nonumber \\
&\overset{(\ref{equation_function_g})}{=} \mathbb{P}_u(u)\, \Big|\frac{dg(u)}{du}\Big|^{-1}. \label{equation_transform_univariate}
\end{align}
The term $|dg(u) / du|^{-1}$ takes into account the change of length during transformation. 

Here, $s$ and $u$ were univariate. If they are multivariate, i.e., $\b{s} , \b{u} \in \mathbb{R}^d$, the above expression becomes:
\begin{align}\label{equation_transform_multivariate}
\mathbb{P}_s(\b{s}) = \mathbb{P}_u(\b{u})\, \Big|\text{det}\big(\frac{\partial g(\b{u})}{\partial \b{u}}\big)\Big|^{-1},
\end{align}
where the term $|\partial g(\b{u}) / \partial \b{u}|^{-1}$ takes into account the change of hyper-volume (volume if dimensionality is three) during transformation.

The derivative $\partial g(\b{u}) / \partial \b{u} \in \mathbb{R}^{d \times d}$ is the Jacobian matrix of $g(\b{u})$ with respect to $\b{u}$. 
Calculation of the determinant of this Jacobian matrix can be difficult and time consuming. 
For this calculation to be simple, we can make the Jacobian matrix a triangular matrix because, according to linear algebra, the determinant of a triangular matrix is the multiplication of its diagonal elements. 
Therefore, we can put restrictions on the Jacobian to become lower triangular. For illustration, assume the dimensionality is three, i.e., $d=3$; then, the Jacobian is:
\begin{align*}
\frac{\partial g(\b{u})}{\partial \b{u}} =
\begin{bmatrix}
\frac{d g_1(\b{u})}{d u_1} & 0 & 0\\
\frac{d g_2(\b{u})}{d u_1} & \frac{d g_2(\b{u})}{d u_2} & 0\\
\frac{d g_3(\b{u})}{d u_1} & \frac{d g_3(\b{u})}{d u_2} & \frac{d g_3(\b{u})}{d u_3}
\end{bmatrix},
\end{align*}
where $u_j$ and $g_j(\b{u})$ denote the $j$-th elements of $\b{u}$ and $g(\b{u})$ respectively. 
In this case, the determinant becomes:
\begin{align}\label{equation_det_diagonal_multiply}
\text{det}\big(\frac{\partial g(\b{u})}{\partial \b{u}}\big) = \prod_{j=1}^d \frac{d g_j(\b{u})}{d u_j},
\end{align}
where $d=3$ in our example.

For having a lower triangular Jacobian matrix, every $g_j(\b{u})$ should be independent of all $\{u_z\}_{z=j+1}^d$ to have zero derivatives with respect to them. In other words, it should be a function of only $\{u_z\}_{z=1}^j$:
\begin{align}\label{equation_gj}
s_j \overset{(\ref{equation_function_g})}{=} g_j(\b{u}) = g_j(u_1, \dots, u_j), \quad \forall j \in \{1, \dots, d\}.
\end{align}
In our three-dimensional example, we should have:
\begin{align*}
& s_1 = g_1(\b{u}) = g_1(u_1), \\
& s_2 = g_2(\b{u}) = g_2(u_1, u_2), \\
& s_3 = g_3(\b{u}) = g_3(u_1, u_2, u_3).
\end{align*}

For the derivative $d g_j(\b{u}) / d u_j$ to be calculated simply in Eq. (\ref{equation_det_diagonal_multiply}), we can choose $g_j(.)$ in Eq. (\ref{equation_gj}) to be affine in terms of $u_j$:
\begin{align}\label{equation_gj_affine}
s_j = g_j(\b{u}) = a_j u_j + b_j,
\end{align}
where $a_j \in \mathbb{R}$ is the slope and $b_j \in \mathbb{R}$ is the intercept for the function $g_j(\b{u})$. 
The function $g_j(\b{u})$ needs to be linear only in terms of $u_j$, so it can be nonlinear and complicated with respect to $u_1, \dots, u_{j-1}$ (note that it should not be a function of $u_{j+1}, \dots, u_d$ according to Eq. (\ref{equation_gj})). 
Therefore, the variables $a_j$ and $b_j$ can be any complicated function of $u_1, \dots, u_{j-1}$:
\begin{equation}\label{equation_a_b}
\begin{aligned}
a_j = a_j(u_1, \dots, u_{j-1}), \\
b_j = b_j(u_1, \dots, u_{j-1}),
\end{aligned}
\end{equation}
so the Eq. (\ref{equation_gj_affine}) becomes:
\begin{align}\label{equation_gj_affine_2}
s_j = g_j(\b{u}) = a_j(u_1, \dots, u_{j-1})\, u_j + b_j(u_1, \dots, u_{j-1}).
\end{align}

As a result, the derivative $d g_j(\b{u}) / d u_j$ in Eq. (\ref{equation_det_diagonal_multiply}) becomes $a_j(u_1, \dots, u_{j-1})$.
The function $g(.)$ needs to be monotone increasing or decreasing to be bijective.
This means that the slope in Eq. (\ref{equation_gj_affine_2}) should not become zero. For ensuring that the slope never becomes zero, we can insert it in an exponential term which never gets zero. Then, Eq. (\ref{equation_gj_affine_2}) becomes:
\begin{align}
s_j &= g_j(\b{u}) \nonumber \\
&= \exp\!\big(a_j(u_1, \dots, u_{j-1})\big)\, u_j + b_j(u_1, \dots, u_{j-1}). \label{equation_gj_affine_3}
\end{align}
In this case, Eq. (\ref{equation_det_diagonal_multiply}) becomes:
\begin{align}\label{equation_det_diagonal_multiply_2}
\text{det}\big(\frac{\partial g(\b{u})}{\partial \b{u}}\big) = \prod_{j=1}^d \exp\!\big(a_j(u_1, \dots, u_{j-1})\big),
\end{align}
which is always a positive number. 

It is noteworthy that the Eq. (\ref{equation_gj_affine_3}) can be solved for $\{u_j\}_{j=1}^d$ iteratively. For example, for three dimensions, it is solved as \cite{durr2020probabilistic}:
\begin{align*}
& u_1 = s_1, \\
& u_2 = \frac{s_2 - b_2(u_1)}{\exp(a_2(u_1))}, \\
& u_3 = \frac{s_3 - b_3(u_1, u_2)}{\exp(a_3(u_1, u_2))}.
\end{align*}

In practice, the parameters of the transformation are obtained by minimizing the negative log likelihood:
\begin{align}\label{equation_optimization_normalizing_flow}
&\text{minimize}\quad - \sum_{i=1}^n \log\big(\mathbb{P}_s(\b{s}_i)\big) \nonumber \\
&~~~~~~~~~~~~~~~~~ \overset{(\ref{equation_transform_multivariate})}{=} - \sum_{i=1}^n \log\Big( \mathbb{P}_u(\b{u}_i)\, \Big|\text{det}\big(\frac{\partial g(\b{u}_i)}{\partial \b{u}}\big)\Big|^{-1} \Big),
\end{align}
where $\b{s}_i$ and $\b{u}_i$ are the $i$-th instance of the sample with size $n$.

\subsection{MADE as an Autoregressive Network}

The explained transformation is the method of normalizing flows \cite{rezende2015variational}. 
It can be implemented using an autoregressive network called MADE \cite{germain2015made}.
The fact that, in Eqs. (\ref{equation_gj}) and (\ref{equation_a_b}), the variables depend only on previous values of the input is the characteristic in autoregressive models which are widely used for time series analysis \cite{shumway2000time}. 
MADE is an autoregressive network which models the function $g(.)$; it takes $\{u_j\}_{j=1}^d$ as input and transforms it to $\{s_j\}_{j=1}^d$. 
Note that this network is not a simple fully connected neural network because it should satisfy Eq. (\ref{equation_gj}). 
Therefore, it uses masks in its layers so that every output $s_j$ depends only on $\{u_z\}_{z=1}^{j}$ and not on $\{u_z\}_{z=j+1}^{d}$. 

Equivalently, the MADE network can output $\{a_j\}_{j=1}^d$ and $\{b_j\}_{j=1}^d$ in Eq. (\ref{equation_gj_affine_3}) rather than $\{s_j\}_{j=1}^d$.
In this case, for satisfying Eq. (\ref{equation_a_b}), the masks of layers should be in a way that every output pair ($a_j$, $b_j$) depends merely on $\{u_z\}_{z=1}^{j-1}$ and not on $\{u_z\}_{z=j}^{d}$. 
The MADE network can be trained by minimizing the log-likelihood to find the parameters.

\begin{figure*}[!t]
\centering
\includegraphics[width=6in]{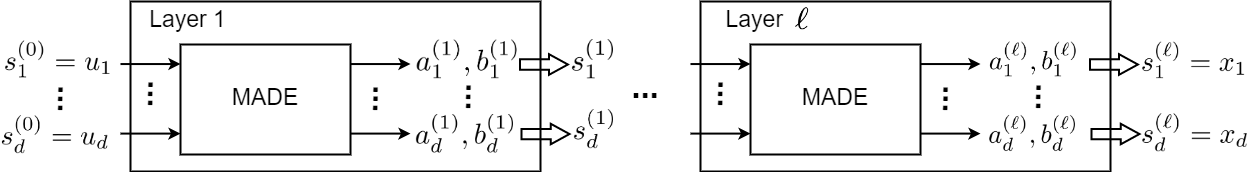}
\caption{MAF model as a stack of $\ell$ layers of MADE modules.}
\label{figure_MAF}
\end{figure*}

\subsection{MAF Network}

Inspired by the deep belief network \cite{hinton2006reducing} which stacks restricted Boltzmann machines, the MAF network is a stack of multiple MADE networks. 
In other words, MAF can be seen as several layers of MADE modules where the output of a MADE module is fed as input to the next MADE module. 

Fig. \ref{figure_MAF} illustrates MAF as a stack of $\ell$ layers of MADE modules.
The latent noise is a $d$-dimensional random variable whose elements are independently and identically distributed as Gaussian distribution, i.e., $u_i \sim \mathcal{N}(0,1)$. 
The first MADE gets $\{u_j\}_{j=1}^d$ as input and outputs $\{a^{(1)}_j\}_{j=1}^d$ and $\{b^{(1)}_j\}_{j=1}^d$. This makes the random variable of the first layer according to Eq. (\ref{equation_gj_affine_3}):
\begin{align*}
s^{(1)}_j = \exp\!\big(a^{(1)}_j(u_1, \dots, u_{j-1})\big)\, u_j + b^{(1)}_j(u_1, \dots, u_{j-1}), 
\end{align*}
$\forall j \in \{1, \dots, d\}$. 
This random variable is then fed as input to the next layer to output $\{a^{(2)}_j\}_{j=1}^d$ and $\{b^{(2)}_j\}_{j=1}^d$:
\begin{align*}
s^{(2)}_j = &\exp\!\big(a^{(2)}_j(s^{(1)}_1, \dots, s^{(1)}_{j-1})\big)\, s^{(1)}_j \\
&+ b^{(2)}_j(s^{(1)}_1, \dots, s^{(1)}_{j-1}), 
\end{align*}
$\forall j \in \{1, \dots, d\}$. 
This goes on until the last layer of MAF. 
In summary, if the initial random noise is denoted to be $\{s^{(0)}_j\}_{j=1}^d := \{u_j\}_{j=1}^d$, the $l$-th layer of MAF performs the following operation:
\begin{equation}
\begin{aligned}
s^{(l)}_j = &\exp\!\big(a^{(l)}_j(s^{(l-1)}_1, \dots, s^{(l-1)}_{j-1})\big)\, s^{(l-1)}_j \\
&+ b^{(l)}_j(s^{(l-1)}_1, \dots, s^{(l-1)}_{j-1}), 
\end{aligned}
\end{equation}
$\forall j \in \{1, \dots, d\}$ and $\forall l \in \{1, \dots, \ell\}$.
The outputs of the last layer determine the generated data, i.e., $\{x_j\}_{j=1}^d := \{s^{(\ell)}_j\}_{j=1}^d$.

The likelihood in MAF is a chain form of the Eq. (\ref{equation_transform_multivariate}):
\begin{align}
&\mathbb{P}_s(\b{s}^{(\ell)}) = \mathbb{P}_u(\b{s}^{(0)})\, \prod_{l=0}^{\ell-1} \Big|\text{det}\big(\frac{\partial g^{(l)}(\b{s}^{(l)})}{\partial \b{s}^{(l)}}\big)\Big|^{-1} \label{equation_transform_MAF} \\
&\implies \log(\mathbb{P}_s(\b{s}^{(\ell)})) = \log(\mathbb{P}_u(\b{s}^{(0)})) \nonumber \\
&~~~~~~~~~~~~~~~~~~~~~~~~~~ - \sum_{l=0}^{\ell-1} \log \Big(\Big|\text{det}\big(\frac{\partial g^{(l)}(\b{s}^{(l)})}{\partial \b{s}^{(l)}}\big)\Big|\Big). \label{equation_transform_MAF_log}
\end{align}
The weights of the MAF network are trained by minimizing the negative log likelihood:
\begin{align}\label{equation_optimization_MAF}
&\text{minimize}\quad - \sum_{i=1}^n \log\big(\mathbb{P}_s(\b{s}_i)\big) \nonumber \\
&~~~~~~~~~~ \overset{(\ref{equation_transform_MAF_log})}{=} - \sum_{i=1}^n \bigg( \log(\mathbb{P}_u(\b{s}_i^{(0)})) \nonumber \\
&~~~~~~~~~~~~~~~~~~~~~~~~~~~ - \sum_{l=0}^{\ell-1} \log \Big(\Big|\text{det}\big(\frac{\partial g^{(l)}(\b{s}_i^{(l)})}{\partial \b{s}^{(l)}}\big)\Big|\Big) \bigg).
\end{align}

\subsection{MAF for Classification}

Similar to the discussion for GMM, MAF can also be used for classification rather than density estimation and data generation out of noise. 
Recall the optimization problem (\ref{equation_maximize_likelihood_times_prior}) in the Bayes' classifier. 
We can use any data generative model, such as MAF, for estimating the likelihood in Eq. (\ref{equation_maximize_likelihood_times_prior}). 
In this case, no matter how complicated the likelihood distribution of every class is and how many modes it has, the MAF model can estimate it. 

Our proposed MAF classifier estimates the likelihood distribution of every class:
\begin{align}
\mathbb{P}(X=\b{x} \,|\, \b{x} \in \mathcal{C}_z) \approx \mathbb{P}_s(\b{s}^{(\ell),z}),
\end{align}
where the superscript $z$ indexes the classes.
Then, the class of a data instance $\b{x}$ is the class which has the maximum posterior among classes:
\begin{align}\label{equation_MAF_classifier}
&\underset{z \in \{1, \dots, |\mathcal{C}|\}}{\text{maximize}}\quad \mathbb{P}(\b{x} \in \mathcal{C}_z) \mathbb{P}_s(\b{s}^{(\ell),z}).
\end{align}

\section{Simulations}\label{section_simulations}

\subsection{Verification by Toy Data}

\begin{figure}[!t]
\centering
\includegraphics[width=3.2in]{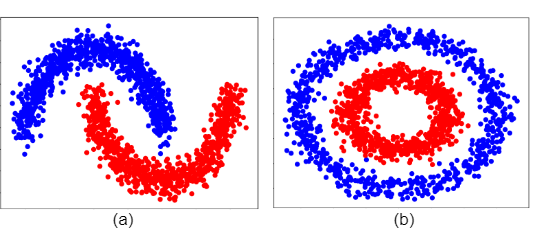}
\caption{The toy datasets: (a) moons data, (b) circles data.}
\label{figure_toy_train_data}
\end{figure}

In this section, we verify the effectiveness of the proposed classifiers using two toy datasets. These datasets, shown in Fig. \ref{figure_toy_train_data}, are the moons and circles datasets which are both nonlinearly separable.

\subsubsection{Performance of GMM Classifier}

\begin{figure}[!t]
\centering
\includegraphics[width=3.2in]{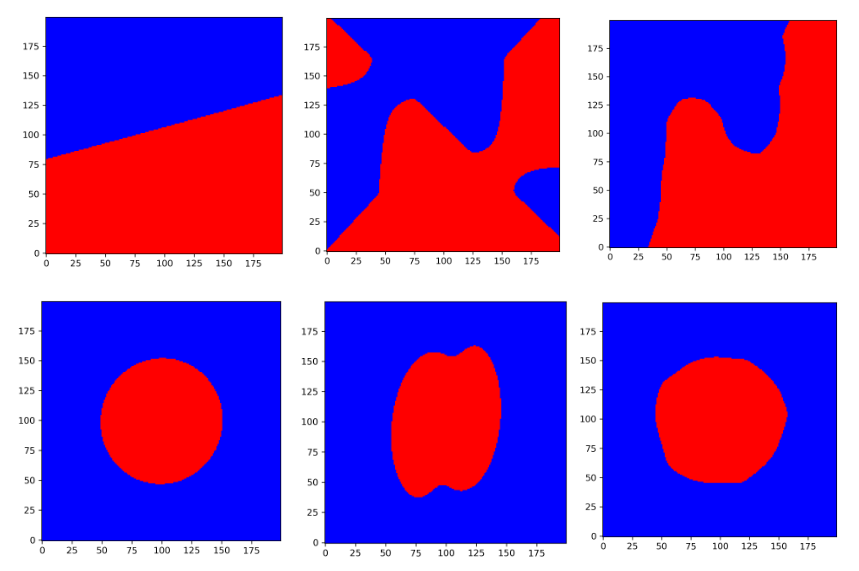}
\caption{The performance of GMM classifier on the toy datasets. The first and second rows correspond to the moons and circles data, respectively. The first to third columns are for GMM with $k=1$, $k=2$, and $k=5$ Gaussians. The colors are for the estimated classes of the test data.}
\label{figure_gmm_toy}
\end{figure}

The performance of the GMM classifier on the toy datasets is illustrated in Fig. \ref{figure_gmm_toy}. 
For the moons data, the GMM classifier with $k=1$ reduces to LDA. This is expected because GMM with $k=1$ fits only one Gaussian, as in LDA, to every class. 
This reduction, however, does not happen for $k=1$ on the circles data because its decision boundary is clearly not linear as in LDA. This also makes sense because one of the classes is surrounded by another class and this makes it perfect for a Gaussian distribution to put a blob around it for classification. This shows that in such cases where a class is surrounded by another class, GMM with $k=1$ does not reduce to LDA. 

Moreover, as Fig. \ref{figure_gmm_toy} shows, increasing the number of Gaussians per class, i.e., $k$, in the GMM classifier can improve the performance because it can model the complicated distributions of the classes with higher degrees of freedom. Although, in some cases, a small $k$ may be sufficient depending on how the classes have been distributed relatively. For example, GMM has performed very well with $k=1$ on the circles data because of the patterns of classes. In general, the value of $k$ should be determined by validation. 

\begin{figure}[!t]
\centering
\includegraphics[width=3.2in]{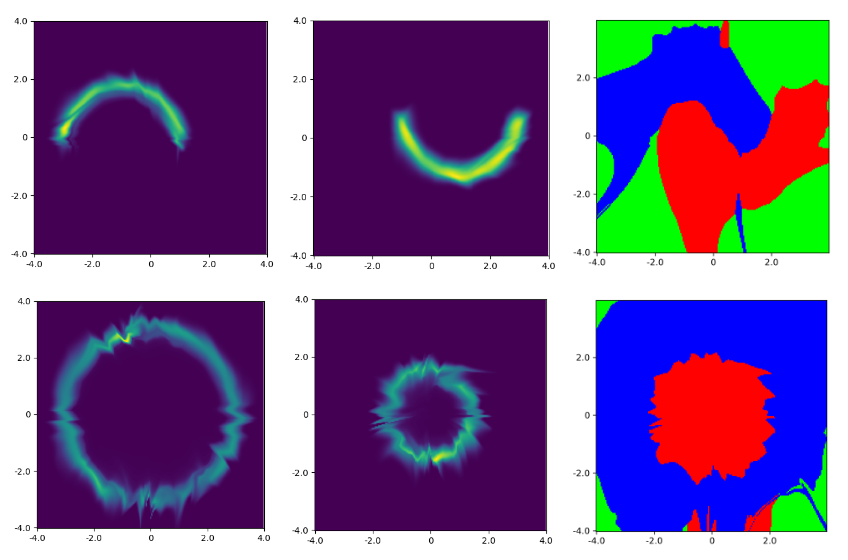}
\caption{The performance of MAF classifier on the toy datasets. The first and second rows correspond to the moons and circles data, respectively. The first and second columns are the best learned latent spaces of training data during training for the first and second classes, respectively. The third column is the estimated classes of the test data. The green color is for the unclassified regions.}
\label{figure_maf_toy}
\end{figure}

\subsubsection{Performance of MAF Classifier}\label{section_toy_MAF}

The performance of the MAF classifier on the toy datasets is depicted in Fig. \ref{figure_maf_toy}. 
This figure shows that the nonlinearly separable classes have been correctly classified by the MAF classifier. 
As the third column in this figure shows, the MAF classifier may leave some regions of the space unclassified. 
This is expected because MAF, as a generative model, is for learning the latent manifold of data and not the whole space. As a result, some outlier regions far away from the training data distribution will have zero probability for all classes. 
For better generalization on outlier
regions, it is recommended to augment data to produce training outliers in the classes.

Having zero or small probabilities in some regions of the space has a benefit, too. The MAF classifier can be used for outlier detection per class, where every data instance may be anomalous (having a zero or very small probability) with respect to a class and normal for other classes. Some data instances and space regions may also be outliers with respect to all classes where their probability is very small for all classes.

Moreover, as shown in the third column of Fig. \ref{figure_maf_toy}, some small regions in the space are classified incorrectly. In these cases, the probability of the estimated class has been a very small positive number. Therefore, it has won the other class whose probability is zero. For resolving this problem, we can filter the estimated probabilities of classes to be higher than a small positive threshold. Then, the mistakenly classified regions in this figure will become unclassified. 

\subsection{Experiments with Real-World Data}

For real-world experiments, we verfied the performance of our proposed density-based classifiers -- MAF and GMM models -- on the publicly available SAHeart (South African Heart Disease Data) \cite{hastie1987non} and Haberman \cite{haberman_dataset} datasets. 
Training, validation, and testing were performed in five-fold cross-validation where the validation data were used for tuning hyperparameters of the MAF and GMM classifiers. 
The tuned hyperparameters of MAF were as follows. For the SAHeart dataset, the optimal number of layers of MAF was $\ell=20$ where every MADE module has eight layers each with $30$ neurons. For the Haberman data, the optimal number of layers of MAF was $\ell=10$ where every MADE module has nine layers each with $5$ neurons. In all experiments, an adaptive learning rate was used with the starting rate $10^{-5}$.
The optimal number of Gaussians in GMM was $k=3$ for both datasets.
The reported test performances are the average test performance across the folds. 

The performance was compared with other baseline classification algorithms including Support Vector Machine (SVM), Logistic Regression (LR), Random Forest (RF), Multi-Layer Perceptron (MLP), LDA, and XGBoost. 
The number of trees in RF was $100$ and the number of hidden layers in MLP was two with $100$ and $50$ neurons, respectively. The rest of hyperparameters of the baseline algorithms were set to the default optimal values in the Python's Scikit-Learn library. 
The codes of the proposed algorithms and the experiments are publicly available at \url{https://github.com/bghojogh/Density-Based-Classifiers}.

\subsubsection{SAHeart Dataset}


The SAHeart dataset \cite{hastie1987non} is widely recognized in the medical research and machine learning areas. It provides valuable information on individuals, particularly males, from a high-risk region for heart disease located in the Western
Cape of South Africa. Researchers utilize this dataset for various types of analyses, including binary classification tasks. It contains eight numerical feature and one categorical feature. 

As Table \ref{table_SAHeart} reports, the proposed MAF model stands out as the best-performing model among the classifiers, with average accuracy 71.86\% and average F1-score 61.00\%.
LR and SVM are strong competitors for the performance on this dataset. 
The performance of our other proposed algorithm, GMM, is almost close to some other classifiers such as MLP.
MAF performs better than GMM because it is a more complex method for joint density estimation.

\setlength{\tabcolsep}{4pt}
\begin{table}[!t]
\centering
\scalebox{1}{
\begin{tabular}{|l|l|l|}
\hline
Model & Average Accuracy & Average F1-Score \\
\hline
GMM & 65.14\% & 42.74\% \\
MAF & {\bf 71.86}\% & {\bf 61.00}\% \\
SVM & 71.20\% & 54.04\% \\
LR & 71.63\% & 54.60\% \\
RF & 67.74\% & 46.60\% \\
MLP & 63.63\% & 43.54\% \\
LDA & 67.74\% & 46.60\% \\
XGBoost & 66.01\% & 46.75\% \\
\hline
\end{tabular}
}
\caption{Results of classification of SAHeart dataset}
\label{table_SAHeart}
\end{table}

\subsubsection{Haberman Dataset}


The Haberman dataset \cite{haberman_dataset} is a well-known benchmark in the fields of medical research and machine learning. It is often used for binary classification tasks, particularly in the context of survival analysis. This dataset contains information about patients who underwent surgery for breast cancer at the University of Chicago's Billings Hospital between 1958 and 1970. The dataset is named after its contributor. 

According to Table \ref{table_Habеrman}, the proposed MAF algorithm achieved the highest accuracy (75.18\%) and F1 score (26.44\%). It demonstrates robustness and strong predictive power for the diagnosis of cardiovascular disease in this dataset. 
MAF, although is complex, works well and is recommended when high accuracy and F1-success are required.
LR had the second best performance for the Haberman dataset. 
Moreover, the proposed GMM algorithm has comparable performance with XGBoost, RF, and LDA. 

It is noteworthy that we also compared our proposed algorithms with baseline algorithms on the well-known diabetes dataset \cite{efron2004least}. MAF and GMM had comparable, but not outperforming, performance on that dataset.


\setlength{\tabcolsep}{4pt}
\begin{table}[!t]
\centering
\scalebox{1}{
\begin{tabular}{|l|l|l|}
\hline
Model & Average Accuracy & Average F1-Score \\
\hline
GMM & 67.98\% & 26.51\% \\
MAF & {\bf 75.18}\% & 26.44\% \\
SVM & 74.19\% & 17.37\% \\
LR & 74.84\% & 24.24\% \\
RF & 68.50\% & 31.46\% \\
MLP & 72.24\% & 27.88\% \\
LDA & 68.50\% & 31.26\% \\
XGBoost & 66.44\% & 33.43\% \\
\hline
\end{tabular}
}
\caption{Results of classification of Haberman dataset}
\label{table_Habеrman}
\end{table}

\section{Conclusion and Future Directions}\label{section_conclusion}

This paper proposed two probabilistic classifiers, GMM and MAF, based on joint density estimation. 
This work opens the research door for other classifiers using joint density estimation, such as Roundtrip GAN \cite{liu2021density}. Density estimation has been widely used in generative models; it can also be used for classification. 
Moreover, the GMM classifier can be modified to use mixture of other distributions based on the data.
Furthermore, as discussed in Section \ref{section_toy_MAF}, the proposed MAF classifier can also be used for outlier detection per class. The advantage of this method is that it can detect anomalies per class where a data instance may be an outlier for a class and normal with respect to another class. 


\section*{Acknowledgement}

The authors thank Sreerekha Rajendran, at Drexel University, for her assistance in experimentation. 












\bibliography{References}
\bibliographystyle{icml2016}

\end{document}